# IDK Cascades: Fast Deep Learning by Learning not to Overthink

**Xin Wang, Yujia Luo, Daniel Crankshaw, Alexey Tumanov, Fisher Yu, Joseph E. Gonzalez**
Department of Electrical Engineering and Computer Sciences
University of California, Berkeley
{xinw, yujialuo, crankshaw, atumanov, fy, jegonzal}@berkeley.edu

## Abstract

Advances in deep learning have led to substantial increases in prediction accuracy but have been accompanied by increases in the cost of rendering predictions. We conjecture that for a majority of real-world inputs, the recent advances in deep learning have created models that effectively "over-think" on simple inputs. In this paper we revisit the classic question of building model cascades that primarily leverage class asymmetry to reduce cost. We introduce the *"I Don't Know"* (IDK) prediction cascades framework, a general framework to systematically compose a set of pre-trained models to accelerate inference without a loss in prediction accuracy. We propose two search based methods for constructing cascades as well as a new cost-aware objective within this framework. The proposed IDK cascade framework can be easily adopted in the existing model serving systems without additional model retraining. We evaluate the proposed techniques on a range of benchmarks to demonstrate the effectiveness of the proposed framework.

## 1 INTRODUCTION

Advances in deep learning have enabled substantial recent progress on challenging machine learning benchmarks. As a consequence, deep learning is being deployed in real-world applications, ranging from automated video surveillance, to voice-powered personal assistants, to self-driving cars. In these applications, accurate predictions must be delivered in *real-time* (e.g, under 200ms) under *heavy query load* (e.g., processing millions of streams) with *limited resources* (e.g., limited GPUs and power).

The need for accurate, low-latency, high-throughput, and low-cost predictions has forced the machine learning community to explore a complex trade-off space spanning model and system design. For example, several researchers have investigated techniques for performing deep learning model compression [1, 2, 3]. However, model compression primarily reduces model memory requirements so as to fit on mobile devices or in other energy-bounded settings. There is a limit to how far compression-based techniques can be pushed to reduce latency at inference time while retaining state-of-the-art accuracy across all inputs.

We conjecture that in the pursuit of improved classification accuracy the machine learning community has developed models that effectively *"overthink"* on an increasing fraction of queries. To support this conjecture we show that while the cost of computing predictions has increased by an order of magnitude over the past 5 years, the accuracy of predictions on a large fraction of the ImageNet 2012 validation images has remained constant (see Fig. 2). This observation suggests that *if* we could distinguish between easy and challenging inputs (e.g., images) and only apply more advanced models when necessary, we could reduce computational costs without impacting accuracy. In this paper we study the design of *prediction cascades* as a mechanism to *exploit this conjecture* by combining fast models with accurate models to increase throughput and reduce mean latency without a loss in accuracy.

Though prediction cascades are well established in the machine learning literature [4, 5, 6, 7], the classic approaches focus on detection tasks and developed cascades for early rejection of negative object or region proposals – leveraging the class asymmetry of detection tasks. In this paper, we revisit the question of how to effectively build model cascades with little training overhead to trade off between prediction accuracy and cost, extending to any multi-class classification task.

We introduce *IDK prediction cascades*, a general framework to accelerate inference without reducing prediction

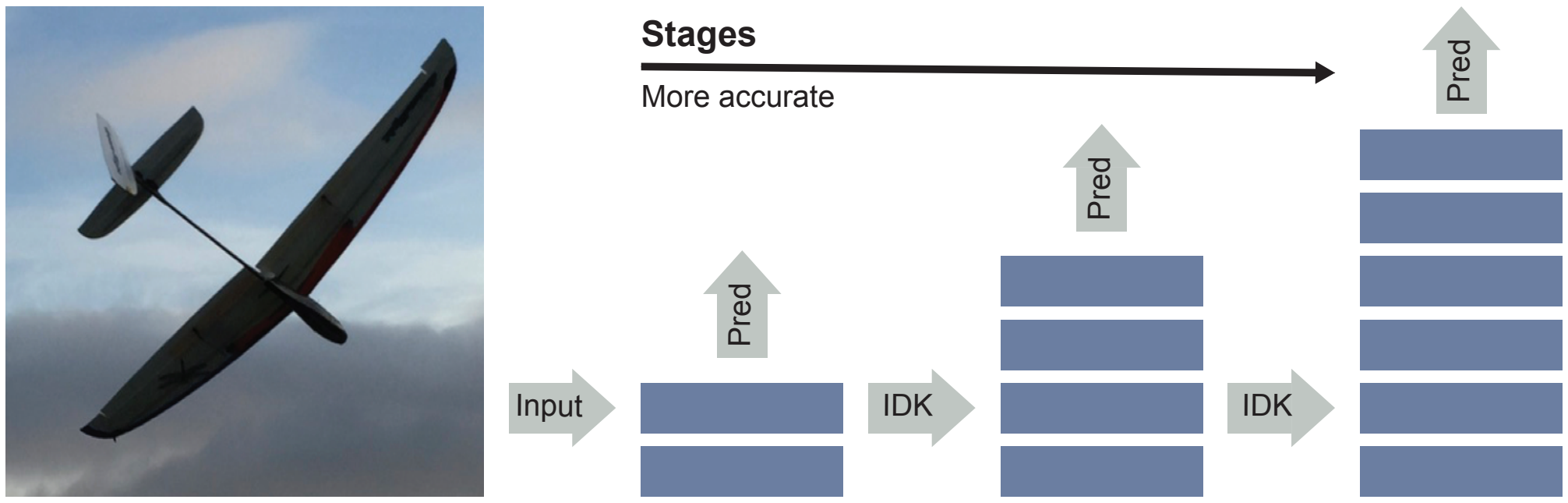

Figure 1: An IDK prediction cascade combines IDK classifiers of increasing accuracy and computational cost such that each will either render a high-accuracy prediction or return IDK passing the input to the next model in the cascade for a more accurate but higher cost prediction

accuracy by composing pre-trained models. IDK prediction cascades (see Fig. 1) are composed of IDK classifiers which are constructed by attaching an augmenting classifier to the existing classifiers, *base models*, enabling the IDK classifiers to predict an auxiliary *"I don't know"* (IDK) class besides the original prediction classes. The augmenting classifiers, which are *light-weighted* (the computational cost is negligible compared to the cost of the base models) and *independent* from the base model architectures, measure the uncertainty of the base model predictions. When an IDK classifier predicts the IDK class the subsequent model in the cascade is invoked. The process is repeated until either a model in the cascade predicts a real class or the end of the cascade is reached at which point the last model must render a prediction. Furthermore, we can introduce a human expert as the last model in an IDK cascade to achieve nearly perfect accuracy while minimizing *the cost* of human intervention.

The base models in the model cascades are treated as black-boxes and thus the proposed framework can be applied to any existing model serving systems [8, 9] with little modification. In addition, the proposed IDK cascade framework model naturally fits the edge-cloud scenario where the fast models can be deployed on edge devices (e.g. Nvidia Drive PX2, Jetson TX2, etc.) while the expensive models are stored in the cloud and are only triggered when the fast model is not certain about a prediction.

To build such model cascade, we need to address the following problems: (1) without retraining the base models or obtaining the base model architecture, what is the best measure available to distinguish the easy and hard examples in the workload? (2) to construct the IDK classifiers that can effectively decide the execution path for the given input while not introducing additional computational overhead, what is the proper objective to balance the computational cost and the overall prediction accuracy of the model cascades?

In this work, we propose two search-based methods for constructing IDK classifiers: *cascading by probability* and *cascading by entropy*. Cascade by probability examines the confidence scores of a model directly to estimate uncertainty. Cascade by entropy leverages well-calibrated class conditional probabilities to estimate model uncertainty. Both techniques then search to find the optimal uncertainty threshold at which to predict the IDK class for each model in the cascade. When the uncertainty in a predicted class exceeds this threshold, the model predicts the IDK class instead. While both search-based methods produce reasonable prediction cascades, neither leverages the cascade design when *training* the augmenting classifier.

As a third approach to constructing IDK classifiers we cast the IDK cascade problem in the context of empirical risk minimization with an additional computational cost term and describe how the objective can be easily incorporated into gradient based learning procedures. The empirical risk minimization based approach allows the IDK classifier to trade-off between *cascade accuracy* and *computational cost* when building the prediction cascade.

We apply all three techniques to the image classification task on ImageNet2012 and CIFAR10 to demonstrate we can reduce computation by 37%, resulting in a 1.6x increase in throughput, while maintaining state-of-the-art accuracy. We conduct a detailed study of the impact of adding the computational-cost term to the objective and show that it is critical for training the augmenting classifiers. Compared to the cascades built with cost-oblivious objectives which cannot usually achieve the desired accuracy, the proposed cost-aware objective better serves the goal of model cascading. Furthermore, we demonstrate that in a real autonomous vehicle setting the IDK cascades framework can be applied in conjunction with human experts to achieve 95% accuracy on driving motion prediction task while requiring human intervention less than 30% of the time.

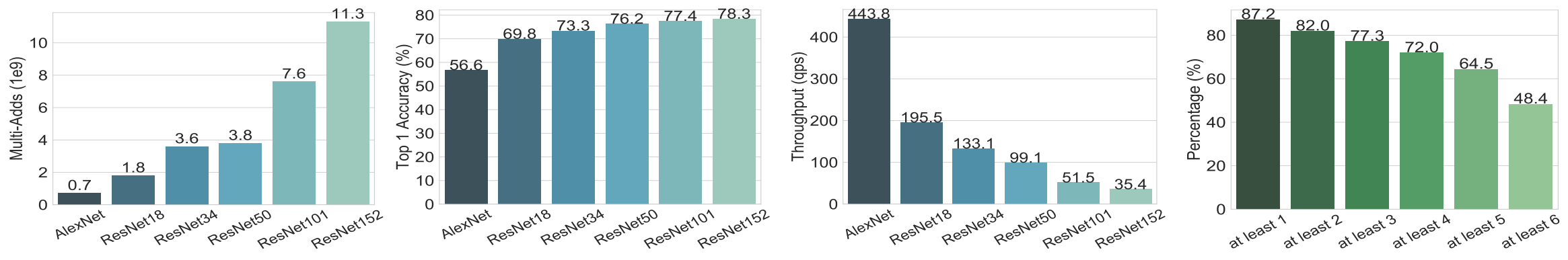

(a) Multi-Adds (b) Accuracy (c) Throughput (d) Easy Examples

Figure 2: *ImageNet Model Statistics*: **(a)** Number of Multi-Adds of the top ImageNet models. *flops* is equivalent to multi-adds in this work and will be used in the following sections. **(b)** The top one prediction accuracy of the top ImageNet models.**(c)** Throughputs (query per second) of the top ImageNet models with batch size = 1. **(d)** The fraction of images correctly labeled by at least K $\in \{1 \dots 6\}$ of the benchmark models

## 2 RELATED WORK

**Compression & Distillation.** Much of the existing work to accelerate predictions from deep neural networks has focused on model compression [3, 2, 1, 10] and distillation [11]. Denton et al. [2] applied low-rank approximations to exploit redundancy in convolution parameters to achieve a factor of two speedup with only 1% reduction in accuracy. Han et al. [10] introduced quantization and Huffman encoding methods to reduce network sizes by orders of magnitude and reduce prediction cost by a factor of 3 to 4. Our work on IDK prediction cascades is complementary to the work on model compression and focuses exclusively on decreasing computation costs. In fact, by coupling model compression with IDK cascades it may be possible to support more aggressive lossy compression techniques. Alternatively, Hinton et al. [11] proposed using soft-targets to transfer knowledge from a costly ensemble to a single model while largely preserving prediction accuracy. Our approach does not require retraining base models and instead focuses on accelerating inference by using more complex models only when necessary. The existing model compression and distillation techniques can be used to construct the fast base models while our framework serves as a bridge to connect models with different levels of complexity and accuracy.

**Cascaded Predictions.** Prediction cascades are a well suited approach to improve prediction performance. Much of the early work on prediction cascades was developed in the context of face detection. While Viola and Jones [5] are credited with introducing the terminology of prediction cascades, prior work by Rowley et al. [4] explored cascading neural networks by combining coarse candidate region detection with high accuracy face detection. More recently, Angelova et al. [6] proposed using deep network cascades and achieved real-time performances on pedestrian detection tasks. Cai et al. [7] also examined cascades for pedestrian detection, proposing a complexity aware term to regularize the cascade objective. While this approach has similarities to the loss function we propose, Cai et al. leverage the cost aware risk to choose an optimal ordering of cascade elements rather than to train a specific classifier. These papers all focus on using cascades for detection tasks, and only use the earlier models in the cascade to reject negative region or object proposals more cheaply. Positive detection (e.g. object identification) can only be made by the final model in the cascade, which is the only model that can predict the full set of classes in the prediction task. Recently, Huang et al. [12] applied the cascading concept by allowing early exiting within the model. Instead of cascading features of a single model, we aim to cascade the trained models (one may not know the model structure or not be able to retrain) in a practical scenario.

**Uncertainty Classes.** The introduction of an IDK class to capture prediction uncertainty has also been studied under other settings. [13, 14]. Trappenberg [13] introduced an *"I don't know"* (IDK) class to learn to identify input spaces with high uncertainty. Khani et al. [14] introduced a *"don't know"* class to enable classifiers to achieve perfect precision when learning semantic mappings. In both cases, the addition of an auxiliary uncertainty class is used to improve prediction accuracy rather than performance. We build on this work by using the IDK class in the construction of cascades to improve performance.

## 3 A MOTIVATING EXAMPLE

In the pursuit of improved accuracy, deep learning models are becoming increasingly expensive to evaluate. To illustrate this trend, in Fig. 2c we plot the throughput for six benchmark models on the ImageNet 2012 datasets. We observe that prediction throughput has decreased by more than an order of magnitude. We expect the trend towards more costly models to continue with improvements in model design and increased adoption of ensemble methods [11]. In contrast, as Fig. 2b shows, the gains in prediction accuracy have increased much more slowly. A result of this trend is that even the cheaper and less advanced models can correctly classify many of the examples.

**Easy Samples.** For many prediction tasks, less accurate models are adequate *most of the time.* For example, a security camera may observe an empty street most of the time and require a more sophisticated model only in the infrequent events that people or objects enter the scene. Even in the standard benchmarks, many of the examples can be correctly classified by older, less advanced models. In Fig. 2d we plot the percentage of images that were correctly classified by an increasing fraction of models. We observe that a large fraction ($\approx 48\%$) of the images are correctly classified by all six of the models, suggesting that these images are perhaps *inherently easier* and may not require the recent substantial increases in model complexity and computational cost.

# 4 IDK PREDICTION CASCADES

We start describing the IDK prediction cascade framework by examining simple two model cascades and then extend these techniques to deeper cascades at the end of this section. We start by formalizing two element cascades for the multiclass prediction problem.

We consider the $k$ class multiclass prediction problem in which we are given two pre-trained models: (1) a fast but less accurate model $m^{\text{fast}}$ and (2) an accurate but more costly model $m^{\text{acc}}$. In addition, we assume that the fast model estimates the class conditional probability:

$$m^{\text{fast}}(x) = \hat{\mathbf{P}}(\text{class label} \mid x). \quad (1)$$

Many multi-class estimators (e.g., DNNs trained using cross entropy) provide class conditional probabilities. In addition, we are given a dataset $\mathcal{D} = \{(x_i, y_i)\}_{i=1}^{n}$ consisting of $n$ labeled data points.

To develop IDK prediction cascades we introduce an additional **augmenting classifier**:

$$h_\alpha(m^{\text{fast}}(x)) \to [0, 1], \quad (2)$$

which evaluates the *distributional output* of $m^{\text{fast}}(x)$ and returns a number between 0 and 1 encoding how *uncertain* the fast model $m^{\text{fast}}(x)$ is about a given prediction. The **IDK classifier** is composed of the base model and the augmenting classifier. For simplicity, we will refer to training the augmenting classifier as training the IDK classifier. In this paper we consider several designs for the IDK classifier:

$$h_\alpha^{\text{prb}}(m^{\text{fast}}(x)) = \mathbb{I}\left[\max_j m^{\text{fast}}(x)_j < \alpha\right] \quad (3)$$

$$h_\alpha^{\text{ent}}(m^{\text{fast}}(x)) = \mathbb{I}\left[\mathbf{H}\left[m^{\text{fast}}(x)\right] > \alpha\right] \quad (4)$$

$$h_\alpha^{\text{cst}}(m^{\text{fast}}(x)) = \sigma\left(\alpha_1 f_{\alpha_2}\left(m^{\text{fast}}(x)\right) + \alpha_0\right), \quad (5)$$

where $\mathbb{I}$ is the indicator function, $\mathbf{H}$ is the *entropy* function:

$$\mathbf{H}[m^{\text{fast}}(x)] = -\sum_{j=1}^{k} m^{\text{fast}}(x)_j \cdot \log m^{\text{fast}}(x)_j, \quad (6)$$

and $f$ is a feature representation of $m^{\text{fast}}(x)$.

While $f$ can be any featurization of the prediction $m^{\text{fast}}(x)$, in this work we focus on the entropy featurization $f = \mathbf{H}$ as this is a natural measure of uncertainty. When using the entropy featurization, the IDK classifier $h^{\text{cst}}$ becomes a differentiable approximation of $h^{\text{ent}}$ enabling direct cost based optimization. In our experimental evaluation, we also evaluate $f_{\alpha_2}(m^{\text{fast}}(x)) = \mathbf{NN}_{\alpha_2}(m^{\text{fast}}(x))$ which recovers a neural feature encoding of $x$ and allows us to assess a neural network based IDK classifier in the context of the differentiable cost based optimization.

Given an IDK classifier $h_\alpha(\cdot)$ we can define a two element IDK prediction cascade as:

$$m^{\text{casc}}(x) = \begin{cases} m^{\text{fast}}(x) & \text{if } h_\alpha(m^{\text{fast}}(x)) <= 0.5 \\ m^{\text{acc}}(x) & \text{otherwise.} \end{cases} \quad (7)$$

Thus, for a given choice of IDK classifier $h_\alpha$ we only need to determine the optimal value for parameter $\alpha$ to ensure maximum accuracy while minimizing the fraction of examples for which the more expensive model is required. In the following, we formalize this objective and describe a set of techniques for choosing the optimal value of $\alpha$.

Given the above definition of an IDK prediction cascade we can define two quantities of interest. We define the accuracy $\mathbf{Acc}(m)$ of a model $m$ as the zero-one prediction accuracy evaluated on our training data $\mathcal{D}$:

$$\mathbf{Acc}(m) = \frac{1}{n}\sum_{i=1}^{n} \mathbb{I}\left[y_i = \arg\max_j m(x_i)_j\right]. \quad (8)$$

We define the IDK rate $\mathbf{IDKRate}(h)$ of an IDK classifier $h$ as the fraction of training examples that are evaluated by the next model in the cascade:

$$\mathbf{IDKRate}(h) = \frac{1}{n}\sum_{i=1}^{n} \mathbb{I}\left[h_\alpha(m^{\text{fast}}(x_i)) > 0.5\right]. \quad (9)$$

Our goal in designing a prediction cascade is then to maintain the accuracy of the more accurate model while minimizing the IDK rate (i.e., the fraction of examples that require the more costly $m^{\text{acc}}$ model). We formalize this goal as:

$$\min_\alpha \; \mathbf{IDKRate}(h_\alpha) \quad (10)$$

$$\text{s.t.:} \quad \mathbf{Acc}(m^{\text{casc}}) \geq (1 - \epsilon)\mathbf{Acc}(m^{\text{acc}}). \quad (11)$$

In the following we describe a set of search procedures for achieving this goal for each of the IDK classifier designs.

### 4.1 BASELINE UNCERTAINTY CASCADES

Similar to [15], as a baseline, we propose using the *confidence scores* (i.e. probability over the predicted class) of $m^{\text{fast}}(x)$ and follow the IDK classifier design ( Eq. 3). The intuition is that if the prediction of $m^{\text{fast}}$ is insufficiently confident then the more accurate classifier is invoked.

A more rigorous measure of prediction uncertainty is the entropy of the class conditional probability. We therefore propose an entropy based IDK classifier in Eq. 4. The entropy based IDK classifier captures the overall uncertainty as well as the certainty within the dominant class.

Due to the indicator functions neither the *cascade by probability* (Eq. 3) nor the *cascade by entropy* functions (Eq. 4) are differentiable. However, because the parameter $\alpha$ is a single scalar, we can apply a simple grid to search procedure to find the optimal value for the threshold $\alpha$.

### 4.2 REGULARIZING FOR PREDICTION COST

The uncertainty based cascades described above adopt a relatively simple IDK classifier formulation and rely on grid search to select the optimal parameters. However, by reframing the cascade objective in the context of regularized empirical risk minimization and defining a differentiable regularized loss we can admit more complex IDK classifiers.

In the framework of empirical risk minimization, we define the objective as the sum of the loss $\mathbf{L}(\cdot,\cdot)$ plus the *computational cost* $\mathbf{C}(\cdot)$ of invoking the cascaded model:

$$J(\alpha) = \frac{1}{n}\sum_{i=1}^{n}[\mathbf{L}(y_i, m_\alpha^{\text{casc}}(x_i)) + \lambda \cdot \mathbf{C}(m_\alpha^{\text{casc}}(x_i))] \tag{12}$$

where $\lambda$ is a hyper parameter which determines the trade-off between the cascade accuracy and the computational cost. Because we are not directly optimizing the fast or accurate models we can adopt the zero-one loss which is compatible with our earlier accuracy goal:

$$\begin{aligned}\mathbf{L}(y_i, m_\alpha^{\text{casc}}(x_i)) = \Big(1 - h_\alpha(m^{\text{fast}}(x))\Big) \cdot \mathbb{I}[y_i, m^{\text{fast}}(x_i)] \\ + h_\alpha(m^{\text{fast}}(x)) \cdot \mathbb{I}[y_i, m^{\text{acc}}(x_i)],\end{aligned} \tag{13}$$

where $\mathbb{I}[y_i, m(x_i)]$ is 1 if $\arg\max_j m(x_i)_j \neq y_i$ and 0 otherwise. The IDK classifier $h_\alpha(m^{\text{fast}}(x))$ governs which loss is incurred. It is worth noting that alternative loss formulations could be used to further fine-tune the underlying fast and accurate model parameters in the cascaded setting.

The cascaded prediction cost $\mathbf{C}(\cdot)$ is defined as:

$$\mathbf{C}(m_\alpha^{\text{casc}}(x_i)) = c^{\text{fast}} + h_\alpha(m^{\text{fast}}(x)) \cdot c^{\text{acc}}, \tag{14}$$

where the computational cost of $m^{\text{fast}}$ and $m^{\text{acc}}$, are denoted by $c^{\text{fast}}$ and $c^{\text{acc}}$ respectively. In practice, the cost of model $m^{\text{fast}}$ and $m^{\text{acc}}$ could be measured in terms of multi-adds, latency, or number of parameters. The formulation of Eq. 14 captures the cascade formulation in which the fast model is always evaluated and the accurate model is evaluated conditioned on the IDK classifier decision.

Combining both prediction loss and computational cost of the IDK cascade, we can now use this *regularized loss* objective function to optimize the IDK prediction cascade with stochastic gradient descent based algorithms. This objective allows us to optimize both prediction precision and overall computational cost in one pass and support more complex parametric IDK classifiers.

### 4.3 BEYOND TWO ELEMENT CASCADES

We can extend the two element cascade to construct deeper cascades by introducing additional IDK classifiers between each model and then either optimizing the IDK classifier parameters in a stage-wise fashion or by jointly optimizing the IDK classifiers using the extended loss function. More precisely, for an $N$-model cascade where $m^j$ is the $j$-th model in the cascade, we define $N-1$ IDK classifiers $h_{\alpha_j}(m^j(x))$. For non-differentiable IDK classifiers with scalar parameters we can apply the grid search procedure in a stage wise fashion starting with the least accurate model. For more complex differentiable IDK classifiers we can define an extended loss:

$$\mathbf{L}(y_i, m_\alpha^{\text{casc}}(x_i)) = \sum_{j=1}^{N}\prod_{q=0}^{j-1} p_q(1-p_j)\mathbb{I}[y_i, m^j(x_i)] \tag{15}$$

where $p_j = h_{\alpha_j}(m^j(x))$ and $p_0 = p_N = 1$. The computational cost function $\mathbf{C}(\cdot)$ is then generalized as

$$\mathbf{C}(m_\alpha^{\text{casc}}(x_i)) = \sum_{j=1}^{N}\prod_{q=0}^{j-1} p_q c^j \tag{16}$$

# 5 EXPERIMENTS

In this section, we evaluate the proposed cascading methods in two scenarios: cascading machine learning models with different computation budgets and collaboration between algorithms and human.

To study the prediction accuracy and cost trade-off under each cascade design, we use standard image classification benchmark tasks and models. We evaluate the cascaded models on ImageNet 2012 [16] and CIFAR-10 [17]. We assess whether the proposed IDK cascade approaches can match the state-of-the-art accuracy while significantly reducing the cost of rendering predictions. We also evaluate the robustness of the proposed framework on CIFAR-10.

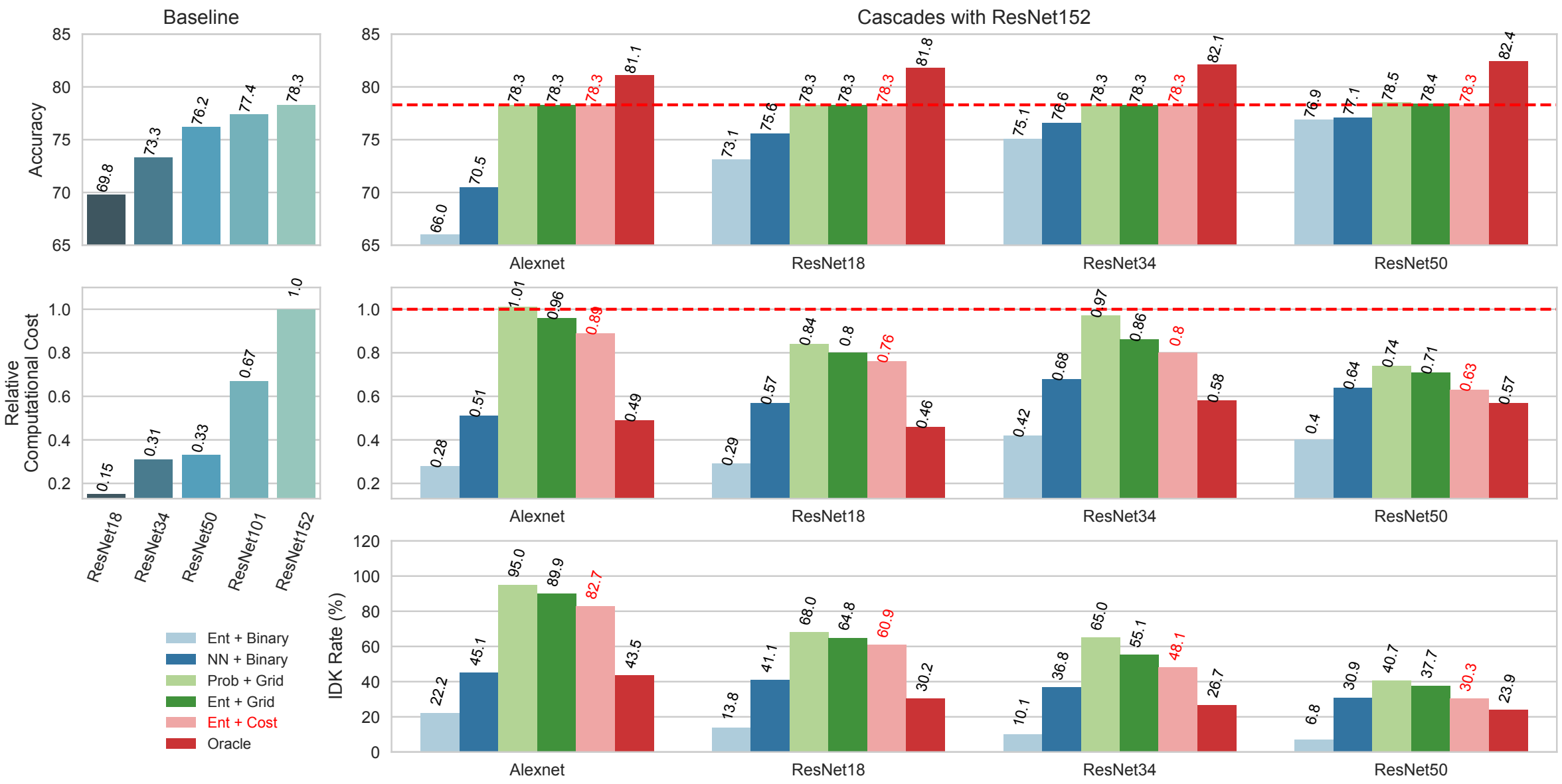

Figure 3: ImageNet Two Model Cascades. **Prob+Grid**: cascade by probability with grid search. **Entropy+Grid**: cascade by entropy with grid search. **NN+Binary**: Neural Network based IDK classifier trained with cost-oblivious cross-entropy loss. **Entropy+Binary**: Entropy based IDK classifier trained with cost-oblivious cross-entropy loss. **Entropy+Cost**: Entropy based IDK classifier with cost-aware objective. **Oracle**: Using ground truth correctness labels as IDK classifier. The comparison of **Prob+Grid**, **Entropy+Grid** and **Entropy+Cost** demonstrates that the proposed cost-aware objective is more effective in constructing model cascades with lower computational costs. The comparison of **NN+Binary**, **Entropy+Binary**, **Entropy+Cost** shows that the vanilla cross-entropy loss commonly used for binary classification leads to a model cascade with lower IDK rate but not achieving the *desired accuracy*

.

To assess how cascades can be used to augment models with human intervention, we evaluate a motion prediction task, a representative of autonomous vehicle workloads [18]. In this human-in-the-loop prediction task, the human serves as the *accurate* model to further improve the accuracy and safety of autonomous driving. The cascade design is used to determine when the fast model can no longer be trusted and human intervention is required (e.g., by taking over steering).

We evaluate each cascade using a range of different metrics. The **accuracy** and **IDK rate** correspond to the accuracy and IDK rate defined in Eq. 8 and Eq. 9 respectively. In the multi-model cascade setting, we measure **IDK Rate** at each level in the cascade. As a measure of computational cost we compute the **average flops** which is the average floating point arithmetic operations required by the model cascade. Finally, as a relative measure of runtime we compute the **relative computational cost** which is computed as $\text{FLOP}_{\text{casc}}/\text{FLOP}_{\text{acc}}$.

## 5.1 IMAGE CLASSIFICATION ON IMAGENET

We first demonstrate that the proposed IDK model cascade framework can preserve the accuracy of the expensive models without a loss while reducing the overall computational cost.

### 5.1.1 Experimental Details

On the ImageNet 2012 dataset we study cascades assembled from pre-trained models including AlexNet [19] and residual networks of various depths including ResNet-18, ResNet-34, ResNet-50, and ResNet-152 [20]. Detailed statistics such as top-1 accuracy, FLOPs, etc of the models are shown in Fig. 2. To train the IDK classifiers, we sample 25.6K training images randomly from the ImageNet 2012 training data and report the cascade accuracy on the entire ImageNet 2012 validation data. In grid search for cascade by probability and entropy, we evaluate 100 different settings of $\alpha$ and select the cascade which has the lowest IDK rate while reaching the desired accuracy. Because 1% reduction in accuracy can translate to a nearly 30% reduction in computational cost on ImageNet (as measured in flops), we set the desired accuracy to be the same as the ResNet-152 (i.e., setting $\epsilon = 0$ in Eq. 11).

For cascade by entropy via cost-aware objective, we set the hyper-parameter $\lambda = 0.04$ across different model combinations and use the actual FLOPs number of each model as the model cost in the objective. We also compare against a cascade constructed using an oracle IDK classifier as a cascade accuracy upper bound which optimally selects between the fast and accurate models. In addition to proposed cascade designs, we include two more IDK cascades constructed by supervised training an IDK classifier of the form in Eq. 5 using the oracle labels with cost-oblivious objectives. We discuss these alternative baselines in more detail in the next section.

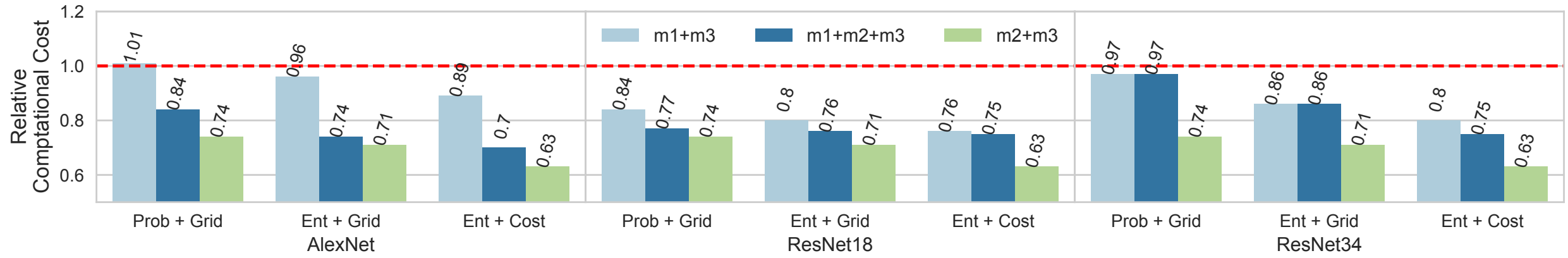

Figure 4: **Three Model Cascade Results.** We consider three element cascades $m_1 \to m_2 \to m_3$ where $m_2$ and $m_3$ are chosen to be the optimal two element cascade consisting of ResNet-50 and ResNet-152 respectively and we evaluate AlexNet, ResNet-18, and ResNet-34 as $m_1$. We also evaluate each of the three IDK cascade designs and corresponding fitting procedure. In all cases, the accuracy is set to match that of ResNet-152 and we therefore only present the computational costs relative to the ResNet-152 model. In general we find that deeper cascades have diminishing returns due to increased evaluation costs

#### 5.1.2 Computation Reduction

Detailed results are shown in Fig. 3. We find the best cascade design employs the Entropy features and regularized cost formulation to combine the ResNet-50 and ResNet-152 models. This cascade is able to reduce prediction costs by 37% while achieving the accuracy of the most computationally expensive model. This is also close to the oracle performance, though it assumes a perfect IDK classifier (i.e. the IDK classifier can distinguish the prediction correctness of the fast model with 100% accuracy and only passing the incorrect predictions to the accurate model). In general we find that our regularized cost based formulation outperforms the other baseline techniques.

#### 5.1.3 Effectiveness of Cost-Aware Objective

In Fig. 3 we also compare the proposed cost based IDK cascade design with two IDK classifiers following the form of Eq. 5 with cost-oblivious cross entropy loss. The training labels are the correctness of the predictions of the fast model evaluated on the ImageNet2012 dataset. We consider two forms of the feature function $f$: entropy based features identical to the cost based cascade and neural network features. The neural network feature function $f_{\alpha_2}(m^{\text{fast}}(x))$ consists of a 7-layer fully connected network with 1024, 1024, 512, 512, 128, and 64 hidden units, ReLU activation functions, and trained using stochastic gradient descent with momentum and batch normalization. In general, we find that these sophisticated baselines are unable to accurately predict the success of the fast model and as a consequence are unable to match the accuracy of the cost based cascade formulation. With the cost-aware objective, the IDK cascades can meet the desired accuracy which shows that the proposed objective is more suitable for building model cascades.

#### 5.1.4 Three Model Cascades

We also investigate three model cascades and the results are shown in Fig. 4. Compared to the two models ResNet-50 + ResNet-152 cascade, adding a faster model like AlexNet, ResNet-18 or ResNet-34 actually *increases* computational cost, because a reasonable fraction of examples will need to pass through all three models in the cascade. However, the three-model cascade tends to reduce the computational cost relative to a two-model cascade including a less accurate model than ResNet-50. Moreover, adding more accurate models within a cascade consistently improves the overall cascade performance.

Table 1: CIFAR Model Details

| Model | % Train Acc | % Test Acc | Flops ($10^7$) |
|---|---|---|---|
| VGG19 | 99.996 | 93.66 | 39.8 |
| ResNet18 | 100.00 | 95.26 | 3.7 |
| DLA-48-B-s | 99.204 | 89.06 | 1.7 |

### 5.2 Robustness Analysis on CIFAR-10

To further analyze the robustness of the IDK prediction cascades, we conduct a set of experiments on the CIFAR-10 datasets. We consider three models: ResNet-18 [20], VGG19 [21] and a recently proposed compact model DLA-48-B-s [22]. Tab. 1 shows details of the models. As we can see from the table, all models overfit the training data with high accuracy close to 100%. We want to study the robustness of the IDK prediction cascades under such extreme case. In this experiment, since VGG19 is less accurate and more costly than ResNet-18, we focus on cascades constructed with DLA-48-B-s and ResNet-18.

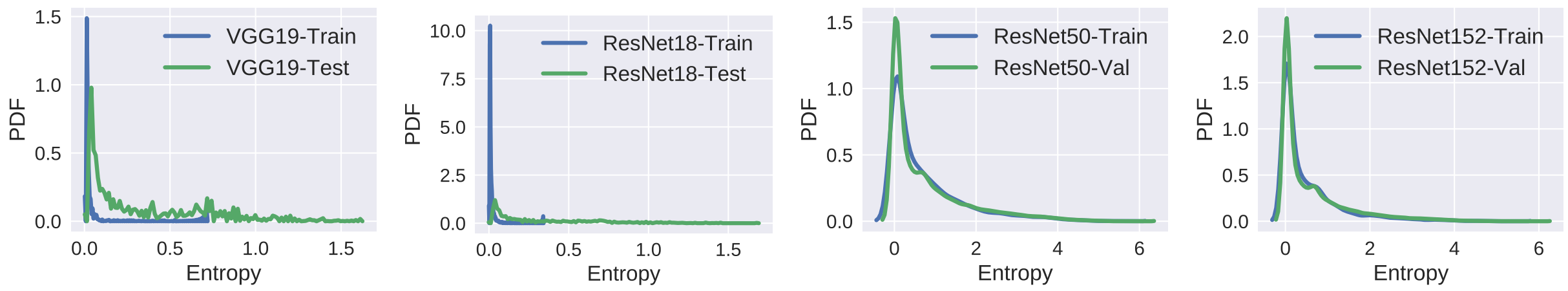

(a) VGG19 on CIFAR (b) ResNet-18 on CIFAR (c) ResNet-50 on ImageNet (d) ResNet-152 on ImageNet

Figure 5: Entropy Distribution. We plot the entropy of the class conditional probability distributions for VGG19 and ResNet-18 on CIFAR-10 and ResNet-50 and ResNet-152 on ImageNet 2012. The VGG model severely overfits the training data and thus can not be used as the base model in the IDK classifier

Table 2: CIFAR Model DLA-48-B-s + ResNet-18 Cascade Results

| Type | Desired Acc (%) | Acc (%) | IDK Rate (%) | Avg Flops ($\times 10^7$) | Relative Computational Cost |
|---|---|---|---|---|---|
| Prob + Grid | 95.26 | 95.23 | 67.5 | 4.198 | 1.135 |
| Entropy + Grid | 95.26 | 95.23 | 51.7 | 3.613 | 0.977 |
| Entropy + Cost | 95.26 | 95.22 | **48.9** | **3.509** | **0.949** |
| Entropy + Grid | 95.16 | 95.16 | 40.1 | 3.184 | 0.861 |
| Entropy + Cost | 95.16 | 95.16 | **40.0** | **3.179** | **0.859** |
| Prob + Grid | 95.05 | 95.07 | 36.5 | 3.051 | 0.824 |
| Entropy + Grid | 95.05 | 95.06 | 33.7 | 2.947 | 0.796 |
| Entropy + Cost | 95.05 | 95.05 | **32.8** | **2.915** | **0.788** |
| Prob + Grid | 94.90 | 94.89 | 29.9 | 2.806 | 0.759 |
| Entropy + Grid | 94.90 | 94.91 | 30.6 | 2.832 | 0.766 |
| Entropy + Cost | 94.90 | 94.91 | **30.5** | **2.827** | **0.764** |

We evaluate the model cascades with four different cascade accuracy goals shown in Tab. 2. We observe cascade by entropy via the cost-aware objective consistently outperforms grid search methods. Also, by admitting a small 0.03% reduction in accuracy, the IDK rate drops substantially from 48.9% to 30.5%. Compared to the single expensive model, the best model cascade reduces computational costs by 24%.

### 5.2.1 Robustness analysis

The proposed IDK classifiers rely on various measures of uncertainty in the class conditional probability distribution and are therefore sensitive to over confidence often as a result of over-fitting. To assess this effect, we evaluate the entropy distribution of the VGG19 and ResNet-18 models which have been trained to near perfect training accuracy (see Tab. 1). We plot the entropy distribution of these models in Figure 5a and 5b on both training and held-out test data and observe that both models substantially over-estimate their confidence on training data when compared with test data. In contrast, the ResNet-50 and ResNet-152 models are much better estimators of prediction uncertainty as seen in Figures 5c and 5d. As a consequence, in settings where the fast model is likely to over-fit it is important to use separate held-out data when training the IDK classifier.

## 5.3 DRIVING CONTROL PREDICTION

We evaluate IDK cascades for autonomous driving and demonstrate that we can achieve nearly perfect accuracy with less than 30% human intervention. In this experiment, we apply the IDK model cascade framework on Berkeley DeepDrive Video dataset, a large scale real driving video dataset [18] containing 2.6 million frames in the training video and 384,599 frames for testing. The driving dataset contains 4 discrete motion states: left turn, right turn, forward and stop. The task is to predict the next vehicle motion given previous video frames. Fig. 6 shows some sample frames in the dataset. We use the *Long-term Recurrent Convolutional Networks* (LRCN) [23] model as $m^{\text{fast}}$ and experiment on a large scale driving video dataset [18]. We consider a human-in-the-loop setting where human serves as the $m^{\text{acc}}$ with 100% accuracy.

We use the same training setting for the proposed cascade approaches as the image classification task and report the results in Tab. 3. The LRCN model has an accuracy of 84.5% [1] and we set the desired accuracy to 100%, 99% and 95%. We find that with only 28.88% human intervention, the cascade model can achieve 95% accuracy which is about 10% more accurate than the base LRCN model. This experiment demonstrates that the model cascade can be easily applied to real applications which are in high demand of low latency and high accuracy.

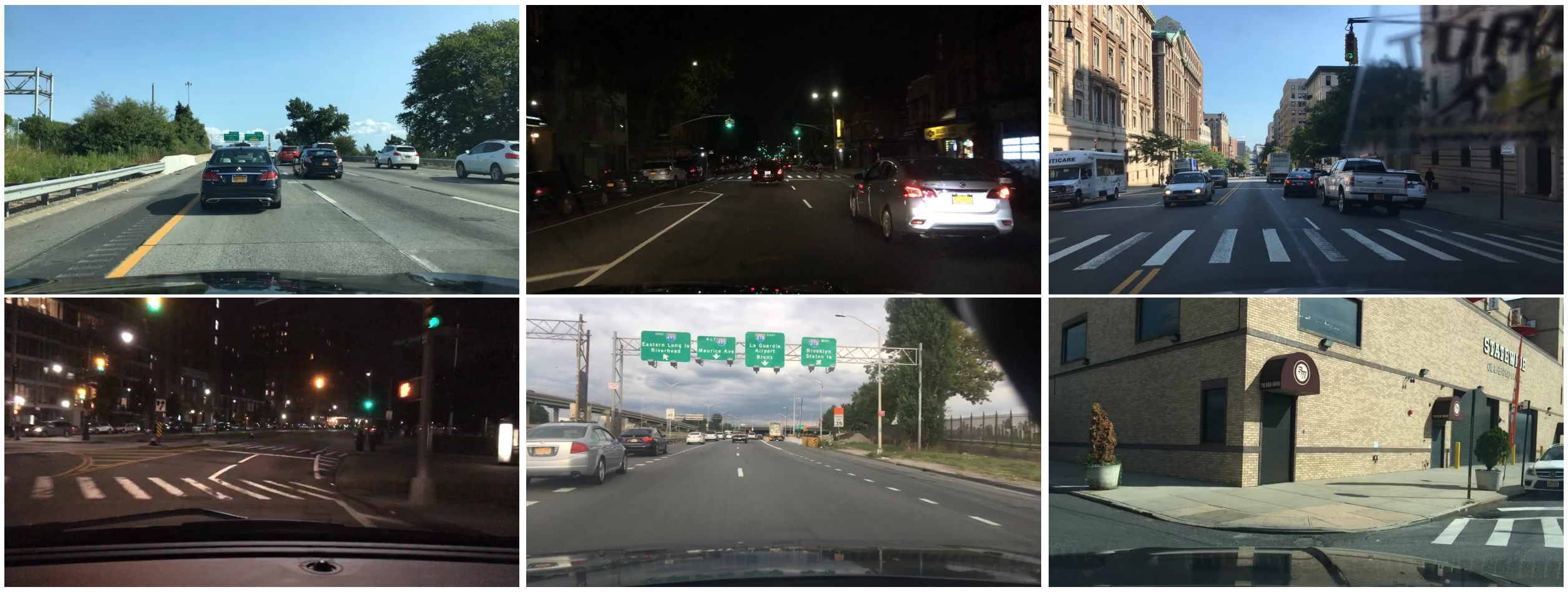

Figure 6: Sample Frames from the Berkeley DeepDrive Video Dataset

Table 3: Driving Model LRCN + Human Expert Cascades

| Type | Desired Acc (%) | Acc (%) | IDK rate (%) |
|---|---|---|---|
| Prob + Grid | 100.0 | 99.9 | 83.91 |
| Ent + Grid | 100.0 | 99.9 | 83.50 |
| Ent + Cost | 100.0 | 99.9 | **80.70** |
| Prob + Grid | 99.0 | 99.2 | 61.72 |
| Ent + Grid | 99.0 | 99.2 | 61.36 |
| Ent + Cost | 99.0 | 99.1 | **59.70** |
| Prob + Grid | 95.0 | 95.4 | 30.08 |
| Ent + Grid | 95.0 | 95.3 | 30.02 |
| Ent + Cost | 95.0 | 95.1 | **28.88** |

## 6 CONCLUSION AND FUTURE WORK

In this paper we revisited the classic idea of prediction cascades to reduce prediction costs. We extended the classic cascade framework focused on binary classifications to multi-class classification setting. We argue that the current deep learning models are "over-thinking" simple inputs in the majority of the real-world applications. Therefore, we aim to learn prediction cascades within the framework of empirical risk minimization and propose a new cost aware loss function, to leverage the accuracy and reduced cost of the IDK cascades.

We focused on build simple cascade with the the pre-trained base models with little training and negligible computation overhead. We tried to answer two questions in this paper: (1) what is a good measure to distinguish the easy and hard examples in the workload without querying much information about the mode itself? We found that the entropy value of the model prediction distribution is a good measure than the vanilla confidence score and can be used as input data to train a light-weighted but effective IDK classifier. (2) How to design the objective function that balances the prediction accuracy and the computation cost? We proposed in this work to use the cost regularized objective which utilizes the actual FLOPs of the base models as the cost measures. Incorporating the cost factor in the objective, we found the model cascade works more effectively than the model cascades with cost-oblivious function.

We also proposed two search based methods *cascade by probability* and *cascade by entropy*, which obtain reasonable performance and require no additional training. We evaluated these techniques on both benchmarks and real-world datasets to show that our approach can successfully identify hard examples in the problem, and substantially reduce the number of invocations of the accurate model with negligible loss in accuracy. We also found that the cost based cascade formulation outperforms uncertainty based techniques.

We believe this work is a first step towards learning to compose models to reduce computational costs. Though not studied in this work, the proposed framework can be easily applied to the existing model serving systems and fit the edge-cloud scenario naturally with little modification. In the future, we would like to explore feature reusing and joint training of the cascade models so that different models can specialize in either easy or hard examples of the given workload.

[1] A slightly reduced accuracy early version was used.

**Acknowledgements**

This research is supported in part by DHS Award HSHQDC-16-3-00083, NSF CISE Expeditions Award CCF-1139158, and gifts from Alibaba, Amazon Web Services, Ant Financial, CapitalOne, Ericsson, GE, Google, Huawei, Intel, IBM, Microsoft, Scotiabank, Splunk and VMware.